\documentclass{article}

\usepackage[english]{babel}

\usepackage[letterpaper,top=2cm,bottom=2cm,left=3cm,right=3cm,marginparwidth=1.75cm]{geometry}

\usepackage{amsmath}
\usepackage{graphicx}
\usepackage{caption}
\usepackage{subcaption}
\usepackage[colorlinks=true, allcolors=blue]{hyperref}

\title{Long-Term Prediction of Natural Video Sequences with Robust Video Predictors}
\author{Luke Ditria, Tom Drummond}

\begin{document}
\maketitle

%===========================================================
\begin{abstract}
Predicting high dimensional video sequences is a curiously difficult problem. The number of possible futures for a given video sequence grows exponentially over time due to uncertainty. This is especially evident when trying to predict complicated natural video scenes from a limited snapshot of the world. The inherent uncertainty accumulates the further into the future you predict making long-term prediction very difficult. In this work we introduce a number of improvements to existing work that aid in creating Robust Video Predictors (RoViPs). We show that with a combination of deep Perceptual and uncertainty-based reconstruction losses we are able to create high quality short-term predictions. Attention-based skip connections are utilised to allow for long range spatial movement of input features to further improve performance. Finally, we show that by simply making the predictor robust to its own prediction errors, it is possible to produce very long, realistic natural video sequences using an iterated single-step prediction task.
\end{abstract}

%===========================================================
\section{Introduction}
As an unsupervised learning task, video prediction gives us the opportunity to learn rich representations of the world that would not be possible using static images alone \cite{huang2018makes}. While looking at a static image alone it is difficult to discern (without direct supervision) between background and foreground objects, or discern individual objects at all. Where does one object start and another end? Only by observing the objects interacting with each other over time can we fully learn their properties. Prediction of video sequences using only video frames is especially difficult as images are a discrete 2D projection of the real world.\\

Many difficulties in frame-to-frame video prediction come from the fact that our system must perform several sub-tasks sequentially in order to create an output that not only looks realistic, but also matches real world data. We argue that these two requirements, predicting a realistic frame and predicting a frame that matches a target frame from the data-set, conflict with each other during training causing poor performance. In order to create plausible predictions in dynamic natural video sequences, we loosen the constraint of matching the target frames exactly. By doing so, we acknowledged that there are some things in the next frame that are not possible to predict at all, let alone with any accuracy.\\
Many works have been able to train deep learning models to predict video sequences into the near future \cite{oprea2020review,oh2015action,lotter2016deep,guen2020disentangling,desai2022next,hsieh2018learning,yu2020efficient,jin2020exploring}. Some works have even been able to produce longer term predictions, but rely on the video sequence having a static background or simple dynamics \cite{villegas2018hierarchical}. Most video prediction models follow a similar design, an auto-encoder style network that takes a sequence of video frames as input. The objective while training is to either predict a single step into the future, or via the use of recurrent models \cite{wang2018eidetic,byeon2018contextvp}, predict multiple steps. There exists other classes of video creation methods, such as video generation, that use adversarial methods to train a video generation model \cite{clark2019adversarial,skorokhodov2022stylegan}. It is even possible to condition generative models on a video sequence to predict future frames \cite{kwon2019predicting,liang2017dual}. In order to attempt to simplify what is already a complicated task, in this work we will be focusing on the end-to-end video prediction task via next frame reconstruction.\\
\begin{figure}[t!]
\centering
\includegraphics[width=120mm]{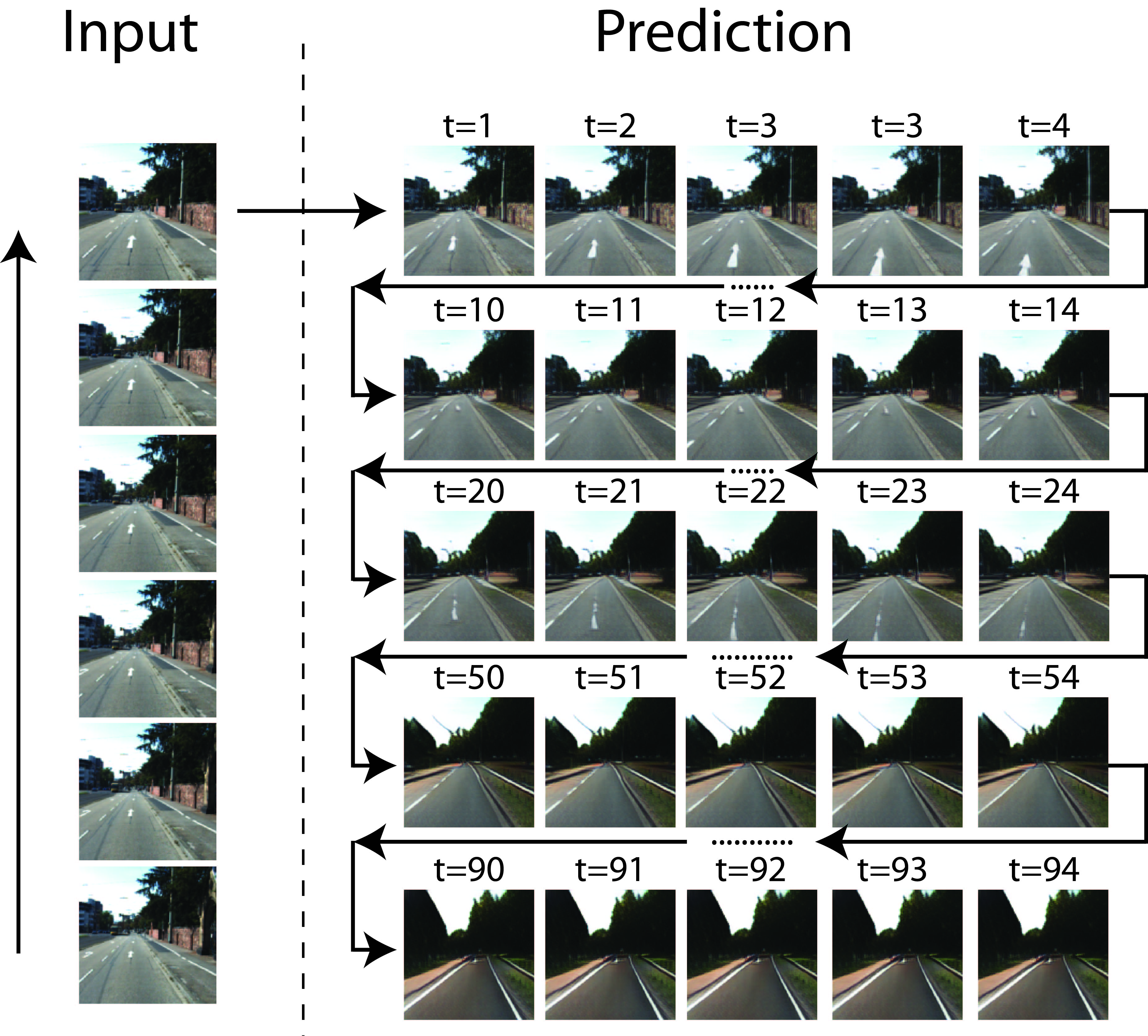}
\caption{Unseen KITTI sequence: From a short input sequence we are able to predict several seconds into the future. Examples videos of predicted sequences are provided as supplementary material}
\label{fig:Skip_Connections}
\end{figure}
For video prediction methods that are trained off of a single-step prediction task, once trained they can be used to perform multi-step prediction by feeding predictions back into the model. These methods rely on their predictions being near perfect in the short term, however for natural video sequences this is very difficult. There will inevitably be errors in the predicted output and these errors, however small, can quickly accumulate and exponentially decay the quality of the output. Part of the issue is that the prediction model was only ever trained on real frames and therefore fully trusts any detected movement in the input sequence. Even nonsensical movements such as those caused by small rendering errors in previous frames are taken into consideration when predicting the next frame and therefore amplified, causing strange artefacts such as warping of the walls of a building.\\
\begin{figure}[b!]
\centering
\includegraphics[width=120mm]{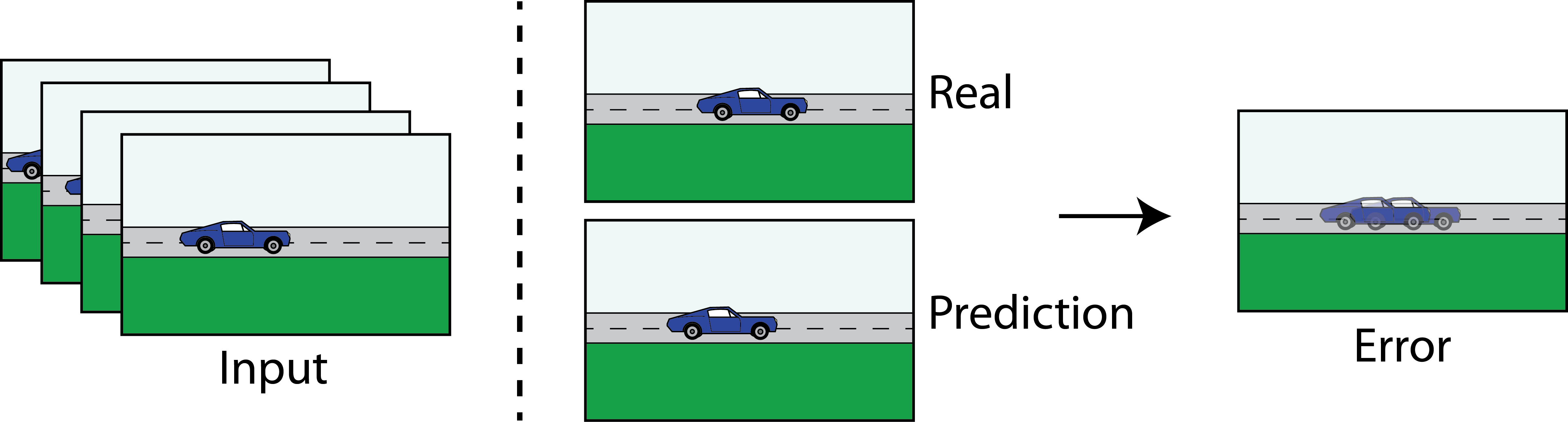}
\caption{Given some information about the world a predictor may come up with a plausible future, however due to a lack of information this prediction may not exactly match what happened. This will result in some error signal, in the case of video prediction this may be a MSE of the pixel values. Unfortunately this error signal does not acknowledge the predictor's lack of information, and as a result the predictor will produce an average of the possible pixel values.}
\label{fig:Car_Example}
\end{figure}
A notable cause of these errors is the uncertainty that is unavoidably introduced when trying to predict in stochastic environments from limited information. For a finite input sequence there exists a large number of possible futures, many of which are equally plausible. Modeling of this sort of uncertainty is usually incorporated into the prediction process by means of variational models \cite{bhattacharyya2018bayesian,donghun2021stochastic,babaeizadeh2017stochastic}. However these methods do not account for the uncertainty that comes about from incomplete or noisy input data \cite{kendall2018multi,byeon2018contextvp}. If we do not take this source of uncertainty into consideration, we will never produce realistic predictions, especially when using pixel based reconstruction losses. To intuit this fact consider the case where our system is trained on a sequence of a car driving along a road (Figure \ref{fig:Car_Example}). If we provide our model with a short sequence of the car moving, we may expect the predictor to carry on the the same trajectory. While this is perfectly logical, perhaps in our actual sequence, right after the sequence we showed the predictor, the car suddenly speeds up. Due to low temporal resolution, this movement might be completely unpredictable, however a naive objective might still punish the predictor for not guessing correctly. In this trivial situation we may remedy the issue by providing more input frames to the model, however the issue still stands that simple pixel reconstruction losses train our models to try and predict exactly what \textit{did} happen, not what \textit{could} happen. In order to account for this, while still trying to match a given target frame, it is possible give our model an additional degree of freedom on its output, a learned uncertainty to discount possible prediction errors \cite{kendall2017uncertainties}. Even after accounting for these various sources of error, it is still incredibly difficult to predict the next frame of a video sequence without minor irregularities. In order to reuse our models predictions to predict further into the future, we must make our predictor robust to its own errors so that it can ignore, or correct, its own mistakes.\\
By taking these points into consideration we propose and demonstrate the following improvements:
\begin{enumerate}
\item An improvement in the quality of predicted next frames by accounting for data uncertainty using a stabilised Gaussian uncertainty loss.
\item When using a Perceptual loss, deeper features from pre-trained classifiers can improve the fidelity of predicted outputs.
\item A simple attention based skip connection that allows the predictor to grab non-local features from previous frames.
\item By feeding the predictor its own predictions during training and allowing it to become robust to its own prediction errors, we can greatly increase the number of time-steps into the future we can predict in a highly dynamic environment.
\end{enumerate}

\section{Related Work}
\subsection{Spatial Information}
When predicting the next frame in a video sequence much of the input frame is redundant information that may not be needed for prediction, but is needed for generating a realistic output frame. We don't want to burden the predictor with trying to preserve high fidelity information while also determining the movement of objects. To achieve this, many frame-to-frame prediction models utilise some sort of skip connection between early layers in the model and deeper ones, similar to a Unet's skip connections \cite{ronneberger2015u}. The issue with this method is that additive or concatenation skip connections only pass information to the same local spatial region. This is fine for image segmentation and similar problems where the object in the output mask is in the same position as the input. However in video prediction we need to account for potentially large shifts in objects between the input and output frames.\\
Existing methods ensure that information in skip connections is mixed so it can be globally accessed by the output layers \cite{shouno2020photo}, though such mixing loses information of the original structure of the image.
Other works aim to solve this issue by training a model to produce, usually several, image transformations or convolutional kernels per input that can be applied to the last input image in a sequence \cite{finn2016unsupervised,babaeizadeh2017stochastic,reda2018sdc}. DNA \cite{finn2016unsupervised} for example, produces a weighting per pixel over a local image patch, CDNA from the same work, does the same with $n$ $5x5$ kernels which encode various translations and applies them to the the last image in the input sequence. By doing so, high fidelity information about objects can be directly moved to its new location in the output. We base our method on spatial attention similar to self attention \cite{zhang2019self} and non-local neural networks \cite{wang2018non}.

\subsection{Uncertainty}
In order to account for the stochasticity of natural video sequences, existing methods usually employ some sort of stochastic sampling method. Commonly used is some form of variational sampling \cite{bhattacharyya2018bayesian,donghun2021stochastic,babaeizadeh2017stochastic} that aims to learn a distribution over the possible futures of high level features. However such methods on their own fail to properly account for data level uncertainty caused by low image resolution (spatial/temporal) or image noise and therefor do not guarantee sharp images. The issue is simple pixel reconstruction losses ($l1$ and $l2$) will still punish plausible, but incorrect, predictions leading to pixel-level uncertainty. To accommodate this, it is possible to predict uncertainty on the output of the model by formulating the pixel reconstruction loss as a negative log-likelihood using a Gaussian distribution to model the error \cite{kendall2017uncertainties,nix1994estimating}. Using this objective the model tries to predict the $\mu$ and $\sigma$ of a Gaussian that will maximise the log-probability of the target value $x$. Using this objective for image reconstruction can lead to sharper outputs as the model can discount pixel errors by producing a high uncertainty $\sigma$. Unfortunately while using this objective (Eq~\ref{eq:Gaussian_Uncertainty}) for prediction, stability issues can arise with the predictor becoming ``over-confident" and producing a very small $\sigma^2$ with an incorrect $\mu$. Some form of regulation over the confidence of the prediction is therefore required. 

\begin{equation} \label{eq:Gaussian_Uncertainty}
    L = \frac{1}{CHW}\sum_i^C\sum_j^H\sum_k^W\frac{(x_{ijk} - \mu_{ijk})^2}{\sigma_{ijk}^2} + \log{(\sigma_{ijk}^2)}~.
\end{equation}

\subsection{Feature Loss}
In order to increase the predicted image quality many methods use feature based losses as well as a simple reconstructive loss. Adversarial losses that utilise a discriminator network are sometimes used \cite{mathieu2015deep,lotter2015unsupervised} but are difficult to train. Another method, that does not require sequential training of additional networks, is to use features extracted from \textit{pre-trained} convolutional classifiers \cite{shouno2020photo,hou2017deep}. These so-called ``Perceptual" losses leverage the feature extracting power of the early convolutional layers to compare the produced image to the target. As these convolutional layers learn to extract important low level features from images, Perceptual losses can be used to greatly improve the image quality of auto-encoder type deep-learning models. 

\begin{figure}[!t]
% \begin{center}
\centering
\begin{subfigure}{0.6\linewidth}
    \centering
    \includegraphics[width=0.9\linewidth]{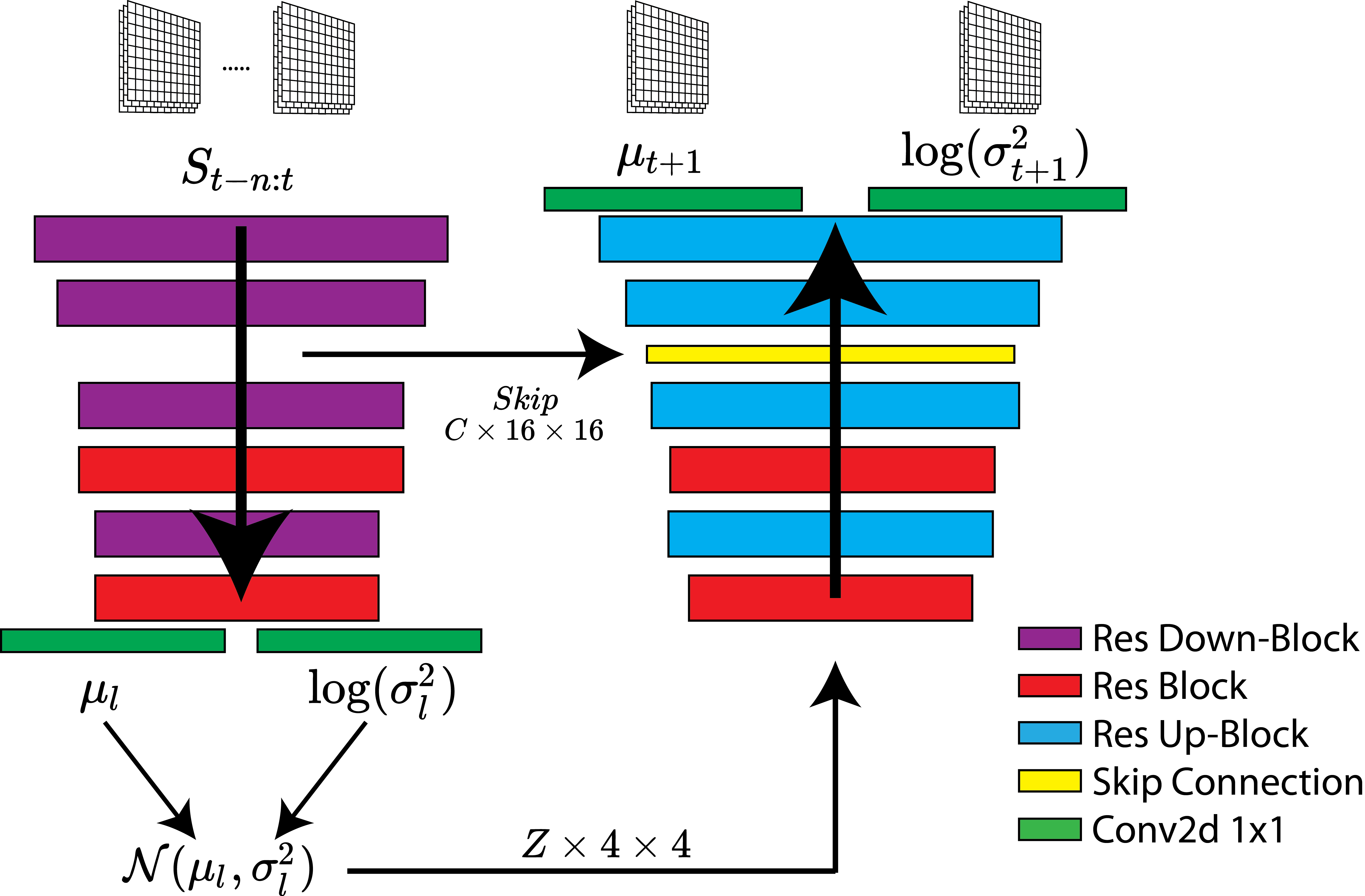}
    \caption{General model architecture.}
    \label{fig:Network}
\end{subfigure}\hfill% 
\begin{subfigure}{0.4\linewidth}
    \centering
    \includegraphics[width=0.8\linewidth]{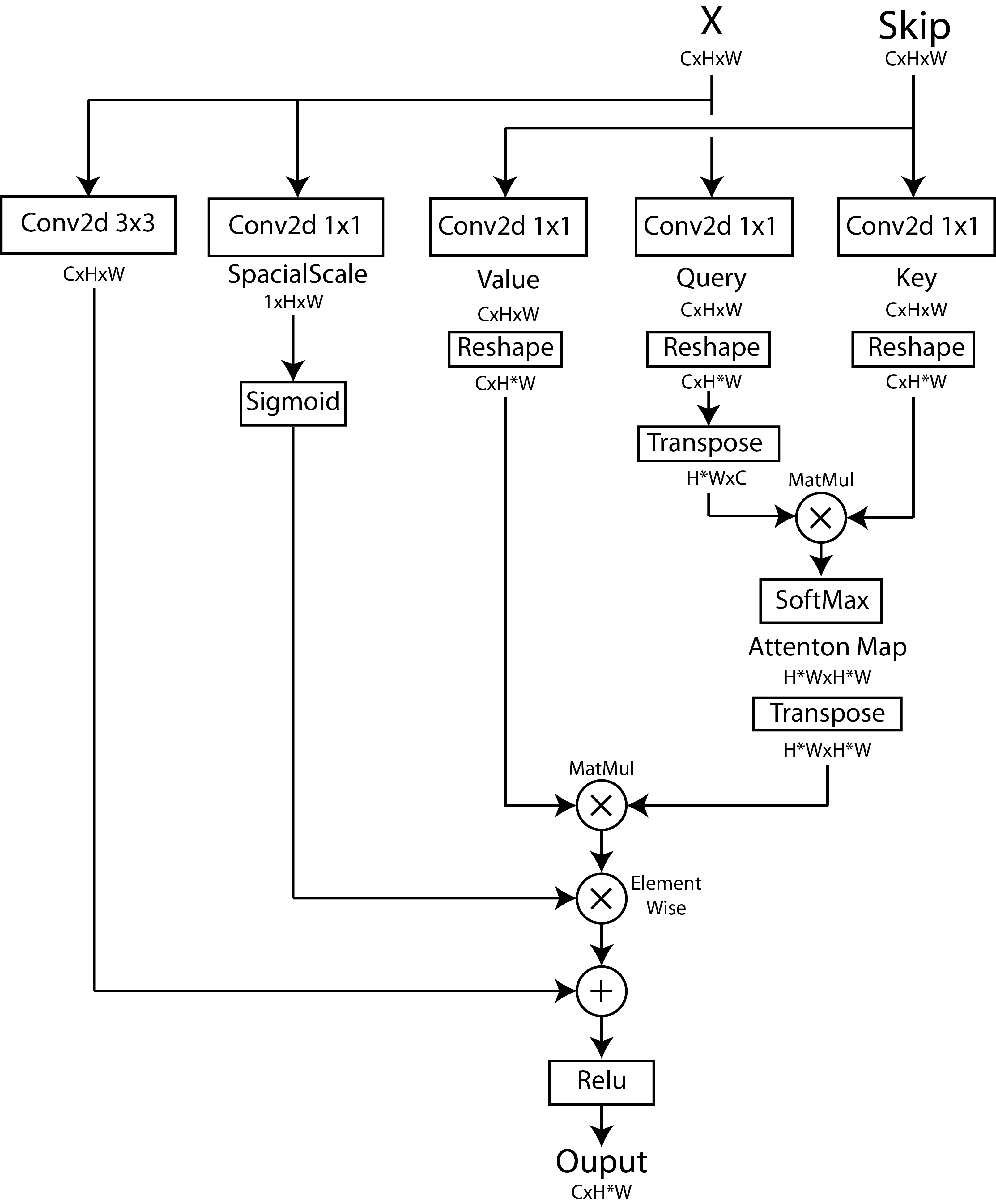}
    \caption{Attention based skip block.}
    \label{fig:Attention}
\end{subfigure}\hfill% 

% \end{center}
   \caption{(a) High level overview of the proposed predictor architecture for 64x64 pixel images. A sequence of n prior frames $S_{t-n}$ to $S_t$ is encoded via the variational encoder. The sampled latent vector is then used by the decoder to select the features from the encoder to produce the $\mu$ and $\sigma$ next frame prediction. (b) The attention based skip block allows the decoder to select features from the encoder to produce the predicted output.}
\end{figure}

\section{Proposed Method}
\subsection{Deep Perceptual loss}
To improve the quality of predicted frames we utilise a deep Perceptual loss. We extract the features from, not only early convolutional layers, but from every convolutional layer within a VGG classifier \cite{simonyan2014very}. To calculate the deep Perceptual loss we simply use the MSE between the raw predicted and target features and average over the layers. By including deep features we shift the predictor's priority from being able to predict low-level features to-high level features. Deep features, by their nature, are robust to small differences in pixel intensity and instead sensitive to-high level structural information. As a result, the predictor will be punished less for not being able to predict the exact value of pixels in the target image. By comparing features of the target and predicted image at all layers, we ensure that predicted objects are consistent in both location and structure.

\subsection{KL Uncertainty}
Noise and the lack of spatial and temporal resolution in images taken from natural video sequences adds pixel-level uncertainty that should be accounted for while predicting. Also, while deep Perceptual losses will force the predictor to produce realistic high-level features that match the target image, lower level features may diverge from the actual next frame. Using deep features can also introduce artifacts into the predicted next frame due to the fact that we are back-propagating through strided convolutions. By using a Gaussian uncertainty-based loss (Eq~\ref{eq:Gaussian_Uncertainty}) we not only allow the predictor to model data uncertainty, but we provide a convenient way for the predictor to ensure that pixel values with high certainty are not changed by the feature loss. If the predictor is certain that its prediction for an output value is correct, it will produce a very low $\sigma^2$; as a result the Gaussian likelihood will be very sensitive to movements in the mean.\\
While this can help retain pixel values that the predictor has high certainty in, the predictor at times may become ``over-confident" producing a very low $\sigma^2$ while it has a high pixel error. As a result, we get large spikes in the loss causing instabilities while training.
To prevent our predictor from becoming over-confident with its predictions, we wish to regulate the maximum certainty of our predictor. This could be done with a simple penalty on $\sigma^2$ however we show that such a penalty emerges if instead of using the maximum log-likelihood of a Gaussian, we instead use the KL divergence of a target distribution and our model's output. \\
Recall the KL divergence between two Gaussians $p=\mathcal{N}(\mu_1, \sigma_1^2)$ and $q=\mathcal{N}(\mu_2, \sigma_2^2$):
\begin{equation}
    KL(p, q) = \log\left(\frac{\sigma_2}{\sigma_1}\right) + \frac{\sigma_1^2 + (\mu_1 - \mu_2)^2}{2\sigma_2^2} - \frac{1}{2}~.
\end{equation}
If we assume that $\sigma^2$ of $p$ is 1 we get:
\begin{equation}
    KL(p, q)= \frac{1}{2}\left( \frac{(\mu_1 - \mu_2)^2}{\sigma_2^2} + \log(\sigma_2^2) + \frac{1}{\sigma_2^2}- 1\right)~.
\end{equation}
We can see that minimising the KL Divergence here is very similar to the Gaussian negative log-likelihood with an additional $1/\sigma^2$ term. Without this term the learned uncertainty $\sigma^2$ is only limited by the error in $\mu$. The inclusion of the term pulls $\sigma^2$ to 1, negating the discounting effect of the uncertainty. We can create a hybrid loss by adding a scaling factor $\alpha$ to the $1/\sigma^2$ term, where $0\leq\alpha\leq1$.  
\begin{equation}
            L = \frac{(\mu_1 - \mu_2)^2}{\sigma_2^2} + \log(\sigma_2^2) + \alpha\frac{1}{\sigma_2^2}~.
\end{equation}
For the case of image prediction this loss is per-element of the output.
\begin{equation} \label{eq:KLLikelihood}
    L = \frac{1}{CHW}\sum_i^C\sum_j^H\sum_k^W\frac{(x_{ijk} - \mu_{ijk})^2}{\sigma_{ijk}^2} + \log{\sigma_{ijk}^2} + \alpha\frac{1}{\sigma_{ijk}^2}~.
\end{equation}

\subsection{Paying Attention}
When predicting the next image in a sequence, it is extremely likely that the output will contain many of the same visual features that are present in the input frames.  These features move due to the motion of the camera and/or objects in the scene. To enable our predictor to use this information, we equip it with a skip-attention layer that is situated near the end of the network and allows features in the output to directly pull visual information from its preferred image location early in the input stage (Fig \ref{fig:Attention}).\\
We generate queries from the decoder's feature map at a chosen resolution and cast attention over feature maps from the encoder at the same resolution [Fig \ref{fig:Network}]. This allows the decoder to access information from any part of the input image in order to generate each part of the predicted output. This can be interpreted intuitively as the encoder learning an image transformation to apply to the input frames and the decoder then applying this transformation and up-sampling. An additive residual connection in the skip block allows the decoder to also create new features not present in the input sequence.

\subsection{Learning to deal with your own mistakes}
Even if we are able to predict the next frame of a sequence accurately small errors will always be present. If we try to feed predictions back into the input of our predictor model, it will treat these errors as real features of the world and include them in future predictions along side any new errors. Therefore, if one tries to perform multi-step prediction by feeding predicted frames back into the predictor many times, these errors will accumulate and the predicted image quality will quickly deteriorate making long video sequence prediction difficult. To counter this problem predicted frames are fed back to the predictor at training time, however we do not back-propagate through multiple time-steps, we simply treat the predicted frame as an ordinary input. In effect, we are using the predictor to augment its own input data thereby training the predictor to be robust to any errors in its own predictions. We perform these steps sequentially for a single sequence, increasing the number of steps into the future that we predict over the course of training. We show that by simply performing this step we can greatly increase the quality of multi-step predictions during inference.

\begin{figure}[t!]
\centering
\includegraphics[width=85mm]{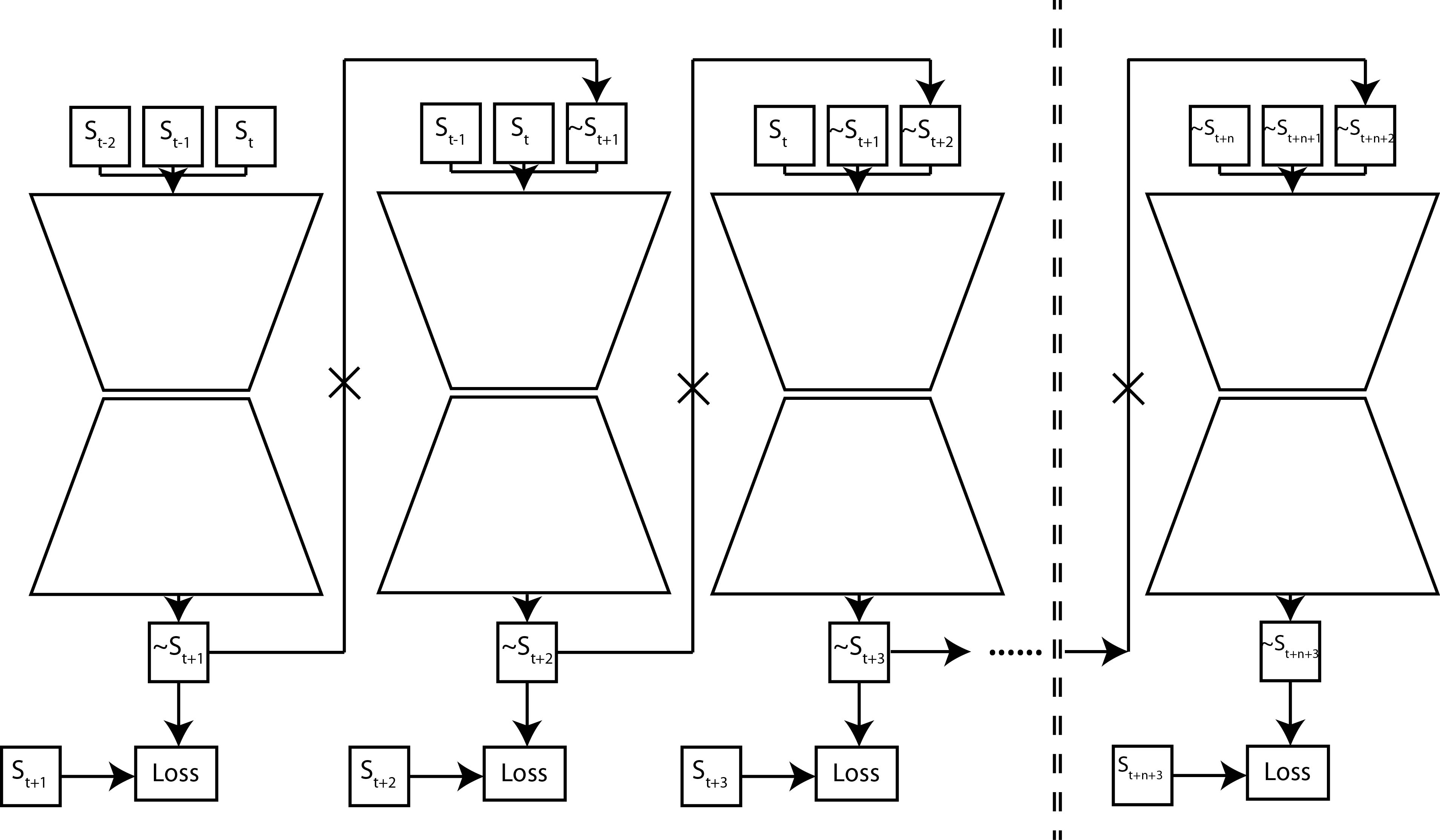}
\caption{Example of the Cycle Training: gradients are \textbf{not} propagated backwards through time, we are simply trying to make the model robust to any errors in its own predictions by replacing real images with its own predictions. This is best done in stages, first with one-step prediction to allow the model to learn from only real inputs. Over the course of training the number of steps is gradually increased until we are performing $n + 1$ steps where $n$ is the number of input frames.}
\label{fig:Cycle}
\end{figure}

\begin{figure}[b!]
\centering
\includegraphics[width=100mm]{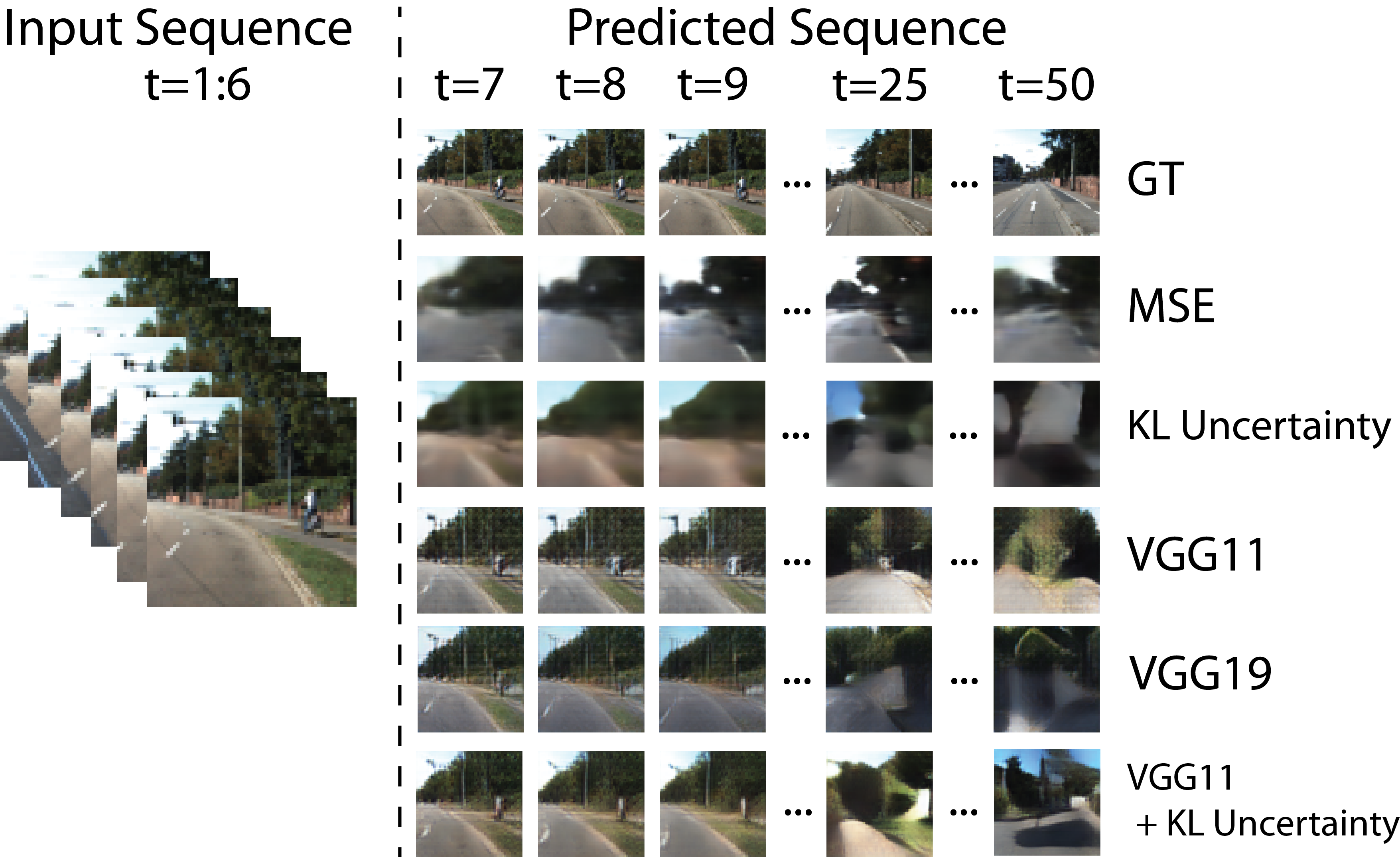}
\caption{Unseen KITTI sequence: Qualitative comparison between identical base architectures trained with various losses. In this input sequence the vehicle is about to turn a corner, the predictor is therefore required to predict the movement of the current objects in the scene \textit{and} generate new features.}
\label{fig:Perceptual_losses}
\end{figure}

\section{Experiments}
\subsection{Training Procedure}
The predictors in the following experiments are trained on 56 driving scenes from the KITTI dataset \cite{geiger2013vision} and test scores are calculated using unseen driving sequences from the Caltech Pedestrian dataset. All examples of predicted sequences were generated from sequences unseen during training. Whole sequences are re-scaled and randomly cropped together to the required resolution. Unless stated otherwise, all experiments are performed at a pixel resolution of 64x64 with the model input being the last 6 frames concatenated along the channel dimension. All of the 64x64 resolution models are trained for a total of 200 epochs. The same general ResNet-VAE based architecture is used for all experiments, adding the skip connections and output uncertainty estimates when necessary (Fig \ref{fig:Network}). During inference the predictor is provided with a short sequence, eg: 6 frames for a model that takes 6 frames as input, and all subsequent predictions are made using previously predicted frames. 

\subsection{Metrics}
\subsubsection{Single-Step Quality}
 Per-pixel MSE can give us a general idea of the similarity between the output frame and the target frame. However, it has been shown that metrics that use learned features, such as LPIPS \cite{zhang2018unreasonable} perform better at providing a human aligned comparison between images than $l2$, PSNR and SSIM.
\subsubsection{Multi-Step Quality}
Determining the quality of predictions multiple time-steps into the future is much more difficult than determining the quality of a single-step prediction. The predicted sequence may diverge from the ``real" sequence while still being a plausible future. We therefore need to look at the quality of the predicted sequence as a whole, to do this we calculate a FVD score \cite{unterthiner2018towards}.

\subsection{Training Objectives}
\subsubsection{Deep Perceptual loss}
We first look at the improvement in prediction quality by training the predictor with features extracted by pre-trained VGG models. We first train a model simply using an MSE loss on the pixel values, we then add the deep Perceptual loss using features from VGG11 and VGG19. We can see that by adding the deep Perceptual loss both the pixel MSE and LPIPS score drop drastically (Table \ref{table:Loss_Metrics}). Visually we can see a large improvement in the quality of the predicted frame and can now make out objects from the input sequence (Figure \ref{fig:Perceptual_losses}). For a 64x64 resolution image, using deeper features from VGG19 does not offer much of an improvement over VGG11, however it should help with larger resolution outputs. Unfortunately, the Perceptual loss also adds artifacts to the output image, and none of these models are able to create realistic predictions very far into the future, with the FVD scores still relatively high. 

\setlength{\tabcolsep}{4pt}
\begin{table}
\centering
\caption{
Quantitative results. Pixel level MSE, PSNR and LPIPS scores are calculated on single step prediction. The FVD score is calculated using a predicted sequence 10 steps into the future with a lower score indicating better performance.
}
\label{table:Loss_Metrics}
\begin{tabular}{lcccc}
\hline\noalign{\smallskip}
Loss & MSE($10^{-3}$) $\downarrow$ & PSNR $\uparrow$ & LPIPS($10^{-2}$) $\downarrow$ & 10 FVD $\downarrow$\\
\noalign{\smallskip}
\hline
\noalign{\smallskip}
MSE & 26.58 & 15.76 & 59.26 & 1962\\
KL & 11.07 & 19.56 & 47.44 & 1675\\
VGG11 + MSE & 18.45 & 0.920 & 29.93 & 1090\\
VGG19 + MSE & 18.78 & 0.763 & 26.95 & 1001\\
VGG11 + KL & 19.52 & 0.901 & 29.83 & 927\\
\hline
\end{tabular}
\end{table}
\setlength{\tabcolsep}{1.4pt}

\subsubsection{Uncertainty}
To more accurately capture the output pixel distribution and to prevent the deep Perceptual loss from introducing unwanted artifacts we allow our model to produce a $\mu$ and uncertainty $\sigma$ for every output pixel value and train using the KL uncertainty instead of MSE. By combining the KL uncertainty loss with the deep Perceptual loss we ensure that, when the predictor is confident, the correct pixel value is produced. When the predictor is uncertain, the deep Perceptual loss ensures that the predictor produces a pixel value that at least helps form consistent high-level features. As a result we can reduce artifacts created by the deep Perceptual loss (Figure \ref{fig:Perceptual_losses}) and create smoother multi-step predictions, lowering the FVD score (Table \ref{table:Loss_Metrics}). 

\subsection{Architecture and Training}

\begin{figure}
\centering
\includegraphics[width=100mm]{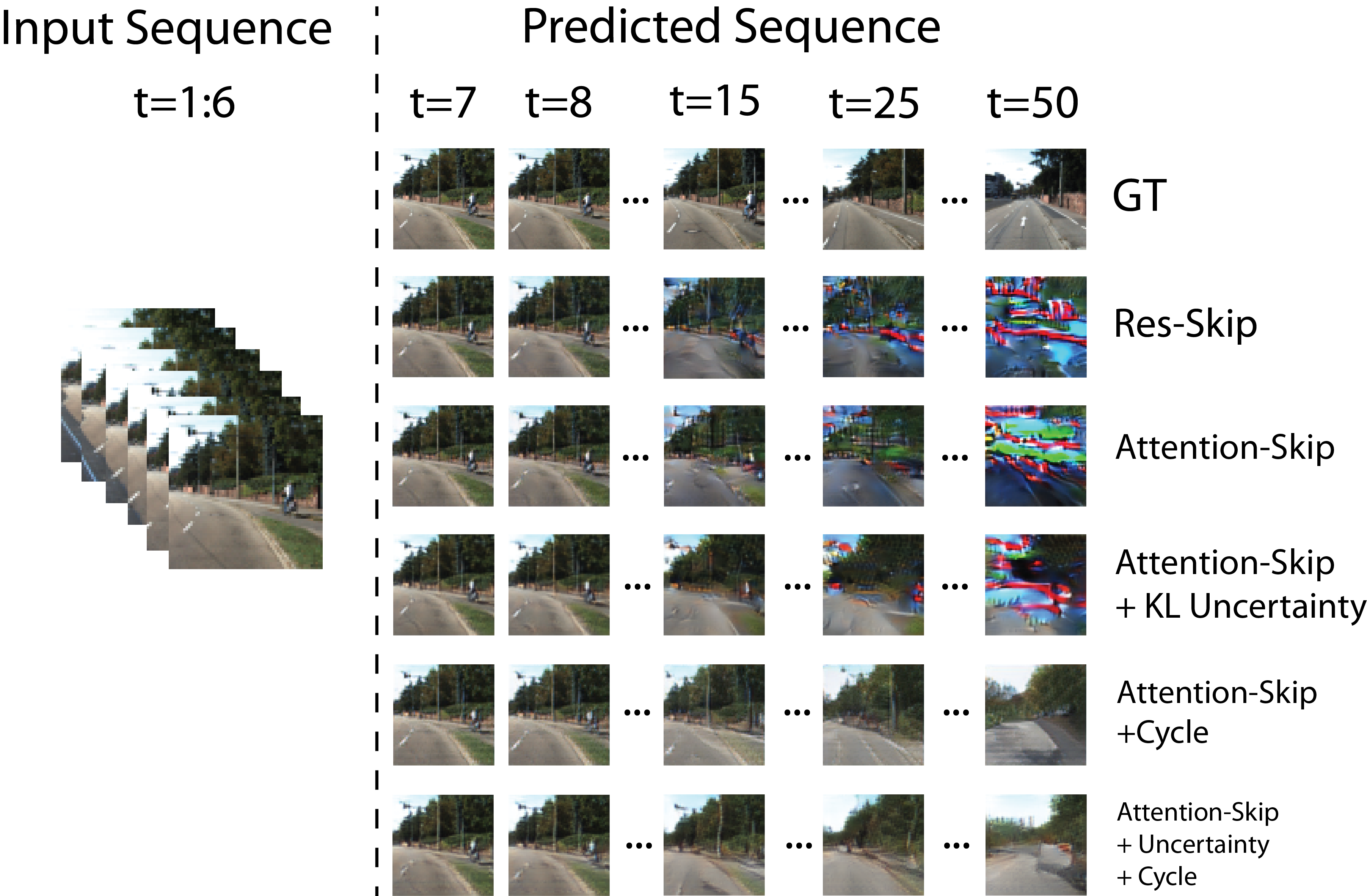}
\caption{Unseen KITTI sequence: Qualitative results from adding skip connections and cycle training.}
\label{fig:Skip_Connections}
\end{figure}

\subsubsection{Skip Attention}
To further improve image quality we add skip connections between the encoder and decoder parts of the predictor. We compare a simple residual skip connection to the attention based skip connection, both placed in the decoder at a 16x16 spacial resolution. We can see that while the residual skip connection provides slightly better single step performance, the attention based skip allows for much higher quality multi-step predictions (Table \ref{table:Loss_Metrics}).

\subsubsection{Cycle Training}
Instead of trying to create a predictor that is able to perfectly learn the distribution of possible next states (which in many cases may be impossible) we instead sacrifice some level of detail in order to predict further into the future. Halfway through training, once the predictor is able to perform a reasonable single step prediction, we start replacing the real images in the model's input for the previously predicted frame. We then get the model to predict the next frame and calculate the losses as if the input were real. As mentioned, we do not back-propagate through the predictor multiple times, instead we simply treat the predicted frame as an augmented version of the real frame. From our results we can see that, while it negatively impacts the image quality of short-term prediction, adding this step greatly improves the quality of long-term prediction. The FVD score of the predicted 50 time-step sequence reduces by more than half in all experiments (Table \ref{table:Metrics}) and qualitative results show that the predictor is able to produce a consistent and realistic sequence after many time-steps (Figure \ref{fig:Skip_Connections}). 

\begin{table}
\centering
\caption{
Quantitative results. PSNR and LPIPS scores are calculated on single step prediction. The FVD score is calculated using a predicted sequence 10 and 50 steps into the future with a lower score indicating better performance.
}
\label{table:Metrics}
\begin{tabular}{llccccc}
\noalign{\smallskip}
\hline
Skip& KL & Cycle & PSNR $\uparrow$ & LPIPS($10^{-2}$) $\downarrow$ & 10 FVD $\downarrow$& 50 FVD $\downarrow$\\
\hline
\hline
Residual & No &  No & 25.76 & 7.9 &  448&  2258\\
Attention & No & No & 24.11 & 10.34 &  371&  1804\\
Attention & Yes & No & 24.69 & 9.25 &  411&  1362\\
\hline
Attention & No & Yes & 21.02 & 17.23&  549&  854\\
Attention & Yes & Yes & 20.34 & 16.32&  471&  700\\
\hline
\end{tabular}
\end{table}
\setlength{\tabcolsep}{1.4pt}

\subsection{Scaling up}
\begin{figure}
\centering
\includegraphics[width=100mm]{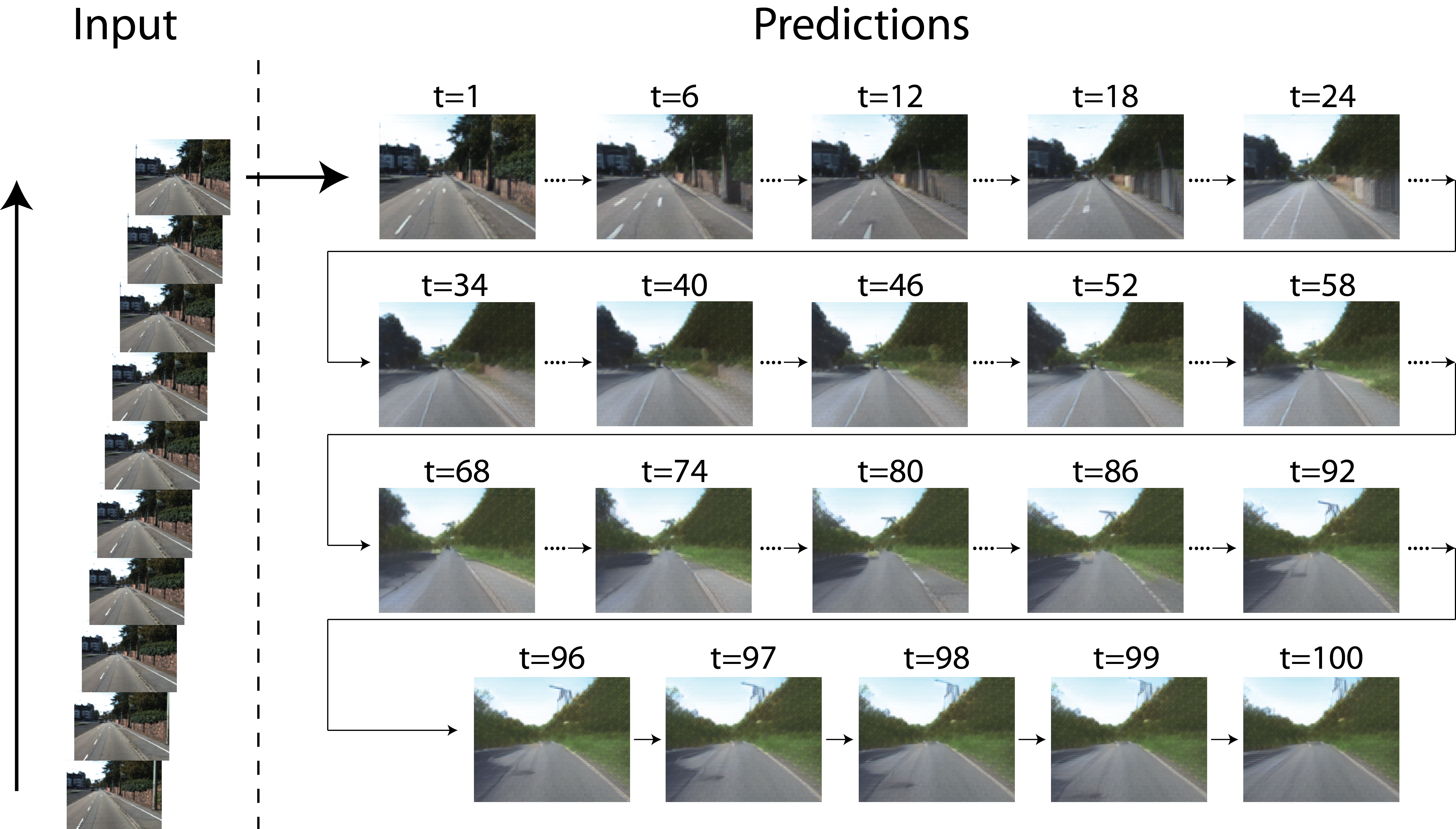}
\caption{Generated 128x128 KITTI driving sequence from unseen test frames.}
\label{fig:KITTI_160}
\end{figure}

To compare to existing methods we scale up our predictor and train with a sequence of 10 128x160 resolution images from KITTI. We incorporate two skip-attention layers, one at a feature size of 16x16 and another at 32x32. We train with a VGG19 deep Perceptual loss and KL uncertainty. Our baseline model without cycle training performs comparatively with existing methods, performing slightly better with the 10 FVD score. When adding cycle training we can see as before that single-step prediction performance decreases but long-term sequence quality dramatically increases (\ref{table:Metrics}). While the level of detail reduces at the start of the sequence, eventually the quality plateaus (Figure \ref{fig:KITTI_160}). This could be the predictor ignoring some level of detail until it is able to confidently predict all features of the sequence. 

\begin{table}
\centering
\caption{
Quantitative results, our RoViP method, with and without cycle training compared against results reported in Chang et al. \cite{chang2021mau}. PSNR and LPIPS scores are calculated on single step prediction. The FVD score is calculated using a predicted sequence 10 and 50 steps into the future with a lower score indicating better performance.
}
\label{table:Metrics}
\begin{tabular}{lccccc}
\hline\noalign{\smallskip}
Method& PSNR $\uparrow$ & LPIPS($10^{-2}$) $\downarrow$ & 10 FVD $\downarrow$& 50 FVD $\downarrow$\\
\hline
\hline
PredNet (ICLR2017) \cite{lotter2016deep} & 27.6 &  9.89 & 2860.8 & -\\
ContextVP (ECCV2018) \cite{byeon2018contextvp} & 28.7 & 9.53 & 2451.6 & -\\
E3D-LSTM (ICLR2019) \cite{wang2018eidetic} & 28.1 & 10.02 & 2311.2 & -\\
Kwon et al. (CVPR2019) \cite{kwon2019predicting} & 29.2 &  8.03 & 1663.2 & -\\
CrevNet (ICLR2020) \cite{yu2020efficient} & 29.3 & 9.11 & 1709.6&  -\\
Jin et al. (CVPR2020) \cite{jin2020exploring} & 29.1 & 8.99 & 1441.1&  -\\
MAU (NeurIPS2021) \cite{chang2021mau} & 30.1 & 8.04 & 1204.0 & -\\
\hline
Ours  & 24.0 &  9.59 & 324.8 & 1918.4\\
Ours + Cycle  (RoViP) & 19.6 &  18.40 & 432.1 & 814.7\\
\hline
\end{tabular}
\end{table}

\section{Conclusions and Future Work}
In this work we have been able to demonstrate that long term video prediction can be achieved by balancing short term fidelity with long term plausibility. Further improvements can be made with the addition of attention based skip connections that allow high resolution features to be moved large spatial distances. Critically, making the predictor robust to its own prediction errors by using predicted frames during training we can greatly extend the number of time-steps into the future we are able create realistic predictions and achieve state-of-the-art long term predictions. In this work we have not incorporated any memory mechanisms, instead focusing on improvements in single step quality and predictor robustness. The addition of long-term memory should help create consistent long term predictions and will be incorporated into future work. We also note that while the predictor is able to predict realistic images far into the future, there is a drop in the level of detail in the sequence before plateauing. Improvements to the model architecture and training regimen may avoid stop this initial drop.

%===========================================================
\bibliographystyle{alpha}
\bibliography{References}

\newcommand{\etalchar}[1]{$^{#1}$}
\begin{thebibliography}{OMGGG{\etalchar{+}}20}

\bibitem[BFE{\etalchar{+}}17]{babaeizadeh2017stochastic}
Mohammad Babaeizadeh, Chelsea Finn, Dumitru Erhan, Roy~H Campbell, and Sergey
  Levine.
\newblock Stochastic variational video prediction.
\newblock {\em arXiv preprint arXiv:1710.11252}, 2017.

\bibitem[BFS18]{bhattacharyya2018bayesian}
Apratim Bhattacharyya, Mario Fritz, and Bernt Schiele.
\newblock Bayesian prediction of future street scenes using synthetic
  likelihoods.
\newblock {\em arXiv preprint arXiv:1810.00746}, 2018.

\bibitem[BWSK18]{byeon2018contextvp}
Wonmin Byeon, Qin Wang, Rupesh~Kumar Srivastava, and Petros Koumoutsakos.
\newblock Contextvp: Fully context-aware video prediction.
\newblock In {\em Proceedings of the European Conference on Computer Vision
  (ECCV)}, pages 753--769, 2018.

\bibitem[CDS19]{clark2019adversarial}
Aidan Clark, Jeff Donahue, and Karen Simonyan.
\newblock Adversarial video generation on complex datasets.
\newblock {\em arXiv preprint arXiv:1907.06571}, 2019.

\bibitem[CZW{\etalchar{+}}21]{chang2021mau}
Zheng Chang, Xinfeng Zhang, Shanshe Wang, Siwei Ma, Yan Ye, Xiang Xinguang, and
  Wen Gao.
\newblock Mau: A motion-aware unit for video prediction and beyond.
\newblock {\em Advances in Neural Information Processing Systems}, 34, 2021.

\bibitem[DJK{\etalchar{+}}21]{donghun2021stochastic}
LEE Donghun, Ingook Jang, Seonghyun Kim, Chanwon Park, and Junhee Park.
\newblock Stochastic video prediction with perceptual loss.
\newblock In {\em NeurIPS 2021 Workshop on Deep Generative Models and
  Downstream Applications}, 2021.

\bibitem[DSC{\etalchar{+}}22]{desai2022next}
Padmashree Desai, C~Sujatha, Saumyajit Chakraborty, Saurav Ansuman, Sanika
  Bhandari, and Sharan Kardiguddi.
\newblock Next frame prediction using convlstm.
\newblock In {\em Journal of Physics: Conference Series}, volume 2161, page
  012024. IOP Publishing, 2022.

\bibitem[FGL16]{finn2016unsupervised}
Chelsea Finn, Ian Goodfellow, and Sergey Levine.
\newblock Unsupervised learning for physical interaction through video
  prediction.
\newblock {\em Advances in neural information processing systems}, 29, 2016.

\bibitem[GLSU13]{geiger2013vision}
Andreas Geiger, Philip Lenz, Christoph Stiller, and Raquel Urtasun.
\newblock Vision meets robotics: The kitti dataset.
\newblock {\em The International Journal of Robotics Research},
  32(11):1231--1237, 2013.

\bibitem[GT20]{guen2020disentangling}
Vincent~Le Guen and Nicolas Thome.
\newblock Disentangling physical dynamics from unknown factors for unsupervised
  video prediction.
\newblock In {\em Proceedings of the IEEE/CVF Conference on Computer Vision and
  Pattern Recognition}, pages 11474--11484, 2020.

\bibitem[HLH{\etalchar{+}}18]{hsieh2018learning}
Jun-Ting Hsieh, Bingbin Liu, De-An Huang, Li~F Fei-Fei, and Juan~Carlos
  Niebles.
\newblock Learning to decompose and disentangle representations for video
  prediction.
\newblock {\em Advances in neural information processing systems}, 31, 2018.

\bibitem[HRM{\etalchar{+}}18]{huang2018makes}
De-An Huang, Vignesh Ramanathan, Dhruv Mahajan, Lorenzo Torresani, Manohar
  Paluri, Li~Fei-Fei, and Juan~Carlos Niebles.
\newblock What makes a video a video: Analyzing temporal information in video
  understanding models and datasets.
\newblock In {\em Proceedings of the IEEE Conference on Computer Vision and
  Pattern Recognition}, pages 7366--7375, 2018.

\bibitem[HSSQ17]{hou2017deep}
Xianxu Hou, Linlin Shen, Ke~Sun, and Guoping Qiu.
\newblock Deep feature consistent variational autoencoder.
\newblock In {\em 2017 IEEE Winter Conference on Applications of Computer
  Vision (WACV)}, pages 1133--1141. IEEE, 2017.

\bibitem[JHT{\etalchar{+}}20]{jin2020exploring}
Beibei Jin, Yu~Hu, Qiankun Tang, Jingyu Niu, Zhiping Shi, Yinhe Han, and
  Xiaowei Li.
\newblock Exploring spatial-temporal multi-frequency analysis for high-fidelity
  and temporal-consistency video prediction.
\newblock In {\em Proceedings of the IEEE/CVF Conference on Computer Vision and
  Pattern Recognition}, pages 4554--4563, 2020.

\bibitem[KG17]{kendall2017uncertainties}
Alex Kendall and Yarin Gal.
\newblock What uncertainties do we need in bayesian deep learning for computer
  vision?
\newblock {\em Advances in neural information processing systems}, 30, 2017.

\bibitem[KGC18]{kendall2018multi}
Alex Kendall, Yarin Gal, and Roberto Cipolla.
\newblock Multi-task learning using uncertainty to weigh losses for scene
  geometry and semantics.
\newblock In {\em Proceedings of the IEEE conference on computer vision and
  pattern recognition}, pages 7482--7491, 2018.

\bibitem[KP19]{kwon2019predicting}
Yong-Hoon Kwon and Min-Gyu Park.
\newblock Predicting future frames using retrospective cycle gan.
\newblock In {\em Proceedings of the IEEE/CVF Conference on Computer Vision and
  Pattern Recognition}, pages 1811--1820, 2019.

\bibitem[LKC15]{lotter2015unsupervised}
William Lotter, Gabriel Kreiman, and David Cox.
\newblock Unsupervised learning of visual structure using predictive generative
  networks.
\newblock {\em arXiv preprint arXiv:1511.06380}, 2015.

\bibitem[LKC16]{lotter2016deep}
William Lotter, Gabriel Kreiman, and David Cox.
\newblock Deep predictive coding networks for video prediction and unsupervised
  learning.
\newblock {\em arXiv preprint arXiv:1605.08104}, 2016.

\bibitem[LLDX17]{liang2017dual}
Xiaodan Liang, Lisa Lee, Wei Dai, and Eric~P Xing.
\newblock Dual motion gan for future-flow embedded video prediction.
\newblock In {\em proceedings of the IEEE international conference on computer
  vision}, pages 1744--1752, 2017.

\bibitem[MCL15]{mathieu2015deep}
Michael Mathieu, Camille Couprie, and Yann LeCun.
\newblock Deep multi-scale video prediction beyond mean square error.
\newblock {\em arXiv preprint arXiv:1511.05440}, 2015.

\bibitem[NW94]{nix1994estimating}
David~A Nix and Andreas~S Weigend.
\newblock Estimating the mean and variance of the target probability
  distribution.
\newblock In {\em Proceedings of 1994 ieee international conference on neural
  networks (ICNN'94)}, volume~1, pages 55--60. IEEE, 1994.

\bibitem[OGL{\etalchar{+}}15]{oh2015action}
Junhyuk Oh, Xiaoxiao Guo, Honglak Lee, Richard~L Lewis, and Satinder Singh.
\newblock Action-conditional video prediction using deep networks in atari
  games.
\newblock {\em Advances in neural information processing systems}, 28, 2015.

\bibitem[OMGGG{\etalchar{+}}20]{oprea2020review}
Sergiu Oprea, Pablo Martinez-Gonzalez, Alberto Garcia-Garcia, John~Alejandro
  Castro-Vargas, Sergio Orts-Escolano, Jose Garcia-Rodriguez, and Antonis
  Argyros.
\newblock A review on deep learning techniques for video prediction.
\newblock {\em IEEE Transactions on Pattern Analysis and Machine Intelligence},
  2020.

\bibitem[RFB15]{ronneberger2015u}
Olaf Ronneberger, Philipp Fischer, and Thomas Brox.
\newblock U-net: Convolutional networks for biomedical image segmentation.
\newblock In {\em International Conference on Medical image computing and
  computer-assisted intervention}, pages 234--241. Springer, 2015.

\bibitem[RLS{\etalchar{+}}18]{reda2018sdc}
Fitsum~A Reda, Guilin Liu, Kevin~J Shih, Robert Kirby, Jon Barker, David
  Tarjan, Andrew Tao, and Bryan Catanzaro.
\newblock Sdc-net: Video prediction using spatially-displaced convolution.
\newblock In {\em Proceedings of the European Conference on Computer Vision
  (ECCV)}, pages 718--733, 2018.

\bibitem[Sho20]{shouno2020photo}
Osamu Shouno.
\newblock Photo-realistic video prediction on natural videos of largely
  changing frames.
\newblock {\em arXiv preprint arXiv:2003.08635}, 2020.

\bibitem[STE22]{skorokhodov2022stylegan}
Ivan Skorokhodov, Sergey Tulyakov, and Mohamed Elhoseiny.
\newblock Stylegan-v: A continuous video generator with the price, image
  quality and perks of stylegan2.
\newblock In {\em Proceedings of the IEEE/CVF Conference on Computer Vision and
  Pattern Recognition}, pages 3626--3636, 2022.

\bibitem[SZ14]{simonyan2014very}
Karen Simonyan and Andrew Zisserman.
\newblock Very deep convolutional networks for large-scale image recognition.
\newblock {\em arXiv preprint arXiv:1409.1556}, 2014.

\bibitem[UvSK{\etalchar{+}}18]{unterthiner2018towards}
Thomas Unterthiner, Sjoerd van Steenkiste, Karol Kurach, Raphael Marinier,
  Marcin Michalski, and Sylvain Gelly.
\newblock Towards accurate generative models of video: A new metric \&
  challenges.
\newblock {\em arXiv preprint arXiv:1812.01717}, 2018.

\bibitem[VEL{\etalchar{+}}18]{villegas2018hierarchical}
Ruben Villegas, Dumitru Erhan, Honglak Lee, et~al.
\newblock Hierarchical long-term video prediction without supervision.
\newblock In {\em International Conference on Machine Learning}, pages
  6038--6046. PMLR, 2018.

\bibitem[WGGH18]{wang2018non}
Xiaolong Wang, Ross Girshick, Abhinav Gupta, and Kaiming He.
\newblock Non-local neural networks.
\newblock In {\em Proceedings of the IEEE conference on computer vision and
  pattern recognition}, pages 7794--7803, 2018.

\bibitem[WJY{\etalchar{+}}18]{wang2018eidetic}
Yunbo Wang, Lu~Jiang, Ming-Hsuan Yang, Li-Jia Li, Mingsheng Long, and
  Li~Fei-Fei.
\newblock Eidetic 3d lstm: A model for video prediction and beyond.
\newblock In {\em International conference on learning representations}, 2018.

\bibitem[YLEF20]{yu2020efficient}
Wei Yu, Yichao Lu, Steve Easterbrook, and Sanja Fidler.
\newblock Efficient and information-preserving future frame prediction and
  beyond.
\newblock 2020.

\bibitem[ZGMO19]{zhang2019self}
Han Zhang, Ian Goodfellow, Dimitris Metaxas, and Augustus Odena.
\newblock Self-attention generative adversarial networks.
\newblock In {\em International conference on machine learning}, pages
  7354--7363. PMLR, 2019.

\bibitem[ZIE{\etalchar{+}}18]{zhang2018unreasonable}
Richard Zhang, Phillip Isola, Alexei~A Efros, Eli Shechtman, and Oliver Wang.
\newblock The unreasonable effectiveness of deep features as a perceptual
  metric.
\newblock In {\em Proceedings of the IEEE conference on computer vision and
  pattern recognition}, pages 586--595, 2018.

\end{thebibliography}

\end{document}